\documentclass{article}
\usepackage[final]{neurips_2021}

\usepackage[utf8]{inputenc}
\usepackage[T1]{fontenc}
\usepackage{url}
\usepackage{booktabs}
\usepackage{amsmath,amsfonts,bm}

\def\eqref#1{equation~\ref{#1}}
\def\1{\bm{1}}

\def\vc{{\bm{c}}}

\def\ve{{\bm{e}}}

\def\vx{{\bm{x}}}

\def\evx{{x}}

\def\mA{{\bm{A}}}

\def\mD{{\bm{D}}}

\DeclareMathAlphabet{\mathsfit}{\encodingdefault}{\sfdefault}{m}{sl}
\SetMathAlphabet{\mathsfit}{bold}{\encodingdefault}{\sfdefault}{bx}{n}

\def\sE{{\mathbb{E}}}

\usepackage{amsmath}
\usepackage{amsfonts}
\usepackage{amssymb}
\usepackage{mathtools}
\usepackage{nicefrac}
\usepackage{microtype}
\usepackage{xcolor}
\usepackage{wrapfig}
\usepackage{caption}
\definecolor{mydarkblue}{rgb}{0,0.08,0.45}
\usepackage[colorlinks=true,
    linkcolor=mydarkblue,
    citecolor=mydarkblue,
    filecolor=mydarkblue,
    urlcolor=mydarkblue]{hyperref}

\title{Baking Symmetry into GFlowNets}

\author{
    Jiangyan Ma\thanks{Correspondence to: \texttt{georgem@stu.pku.edu.cn}}\\
    Peking University\\
    \And
    Emmanuel Bengio\\
    Recursion Pharmaceuticals\\
    \AND
    Yoshua Bengio\\
    Mila, Université de Montréal, CIFAR\\
    \And
    Dinghuai Zhang\\
    Mila, Université de Montréal\\
}

\begin{document}

\maketitle

\begin{abstract}
    GFlowNets have exhibited promising performance in generating diverse
    candidates with high rewards. These networks generate objects
    incrementally and aim to learn a policy that assigns  probability of
    sampling objects in proportion to rewards. However, the current training
    pipelines of GFlowNets do not consider the presence of isomorphic actions,
    which are actions resulting in symmetric or isomorphic states. This lack of
    symmetry increases the amount of samples required for training GFlowNets and
    can result in inefficient and potentially incorrect flow functions. As a
    consequence, the reward and diversity of the generated objects decrease. In
    this study, our objective is to integrate symmetries into GFlowNets by
    identifying equivalent actions during the generation process. Experimental
    results using synthetic data demonstrate the promising performance of our
    proposed approaches.
\end{abstract}

\section{Introduction}

Generative Flow Networks (GFlowNets, \citet{gflownet}) have been proposed as a
method to generate a wide range of high-quality candidates. By learning a stochastic
policy $\pi$ for generating discrete objects $x$, GFlowNets iteratively add simple
building blocks to partial objects, resulting in diverse and high-scoring candidates.
These networks have shown promising performance in various tasks such as diverse
molecule generation, active learning, biological sequence design, graph
combinatorial optimization, and probabilistic learning \citep{gflownet,
biological-sequence-design,let-the-flows-tell,probabilistic-modeling,
bayesian-structure-learning}.

However, prior works mostly neglect the internal symmetries within the generation
process, leading to redundant data representations. Recent theoretical findings
highlight the potential for improved sample complexity by incorporating data
symmetry \citep{sample-complexity-gain}, and several approximation strategies for
invariance have been proposed in the context of Graph Neural Networks (GNNs)
\citep{relational-pooling,janossy-pooling,forcenet,rotation-invariant,
frame-averaging,laplacian-canonization}. Unfortunately, the current training
pipelines for GFlowNets do not consider the existence of symmetric states and
actions. This oversight could result in increased sample complexity and potentially
incorrect flow probabilities, ultimately impacting the diversity and average reward
of the generated objects.

To address this issue, we present two approaches in the GFlowNet training process
that enforce invariance to isomorphic states and actions. When faced with isomorphic
states, we suggest using canonization techniques to reduce the states (partial
objects) to their canonical form, thereby reducing the size of the state space. For
the graph generation process, isomorphic actions are defined as actions that lead
to isomorphic preceding graphs, but these actions lack an efficient canonical form. In
this scenario, we propose the use of handcrafted positional encodings (PE) to
identify isomorphic actions efficiently while maintaining expressive power. Our
synthetic experiments demonstrate the effectiveness of these proposed approaches.

\section{Background}

We will examine the framework proposed by \citet{gflownet}, which involves a pointed
directed acyclic graph (DAG) denoted as $(\mathcal{S},\mathbb{A})$. In this setting,
there is a designated initial state, $\mathcal{S}$ consists of a finite set of vertices
called \textit{states}, and $\mathbb{A}\subset\mathcal{S}\times\mathcal{S}$ represents
a set of directed edges known as \textit{actions}. If an action $\mathbf{s}\to\mathbf
{s}'$ exists, we refer to $\mathbf{s}$ as the \textit{parent} and $\mathbf{s}'$ as the
\textit{child}. Additionally, there is precisely one state that has no incoming edge,
identified as the \textit{initial state} $\mathbf{s}_0\in\mathcal{S}$. States without
outgoing edges are referred to as \textit{terminating}. We denote the set of
terminating states as $\mathcal{X}$. A \textit{complete trajectory} is a sequence
$\tau=(\mathbf{s}_0\to\dots\to\mathbf{s}_n)$ such that each $\mathbf{s}_i\to\mathbf{s}
_{i+1}$ is an action and $\mathbf{s}_n\in\mathcal{X}$. We represent the set of complete
trajectories as $\mathcal{T}$, and $\mathbf{x}_\tau$ indicates the last state of a
complete trajectory $\tau$.

GFlowNets belong to a category of models that amortize the cost of sampling from an
intractable target distribution over $\mathcal{X}$. These models accomplish this by
learning a functional approximation of the target distribution using its unnormalized
density or reward function denoted as $R\colon\mathcal{X}\to\mathbb{R}^+$.
\citet{gflownet} defines a trajectory flow $F\colon\mathcal{T}\to\mathbb{R}_{\geq 0}$.
We can define a state flow $F(\mathbf{s})=\sum_{\tau\ni\mathbf{s}}F(\tau)$ for any
state $\mathbf{s}$ and an edge flow $F(\mathbf{s}\to\mathbf{s}')=\sum_{\tau\ni\mathbf
{s}\to\mathbf{s}'}F(\tau)$ for any edge $\mathbf{s}\to\mathbf{s}'$. The trajectory flow
induces a probability measure $P(\tau)=\frac{F(\tau)}{Z}$, where $Z=\sum_{\tau\in
\mathcal{T}}$ represents the total flow. Furthermore, we define the forward policy
$P_F(\mathbf{s}'|\mathbf{s})=\frac{F(\mathbf{s}\to\mathbf{s}')}{F(\mathbf{s})}$ and
the backward policy $P_B(\mathbf{s}|\mathbf{s}')=\frac{F(\mathbf{s}\to\mathbf{s}')}
{F(\mathbf{s}')}$. In this context, the flows can be likened to the amount of water
flowing through edges (similar to pipes) or states (resembling tees connecting pipes)
\citep{tb}. $R(\mathbf{x})$ represents the amount of water passing through the terminal
state $\mathbf{x}$, while $P_F(\mathbf{s}'|\mathbf{s})$ corresponds to the relative
quantity of water flowing in edges originating from $\mathbf{s}$.

GFlowNet has proven effective for generating both diverse and high-quality data in structured domains, including molecule optimization~\citep{gflownet,jain2022multiobjective}, causal discovery~\citep{bayesian-structure-learning,deleu2023joint}, combinatorial optimization~\citep{zhang2022schedulinggflownets,let-the-flows-tell}, and neural network structure learning~\citep{Liu2022GFlowOutDW}. Much connection has been drawn between GFlowNets and probabilistic modeling~\citep{probabilistic-modeling,malkin2022gfnhvi,zhang2022unifying,zhou2024phylogfn,zhang2024diffusion,zhang2024improving} and control methods~\citep{pan2022gafn,pan2023better,pan2023stochastic,zhang2024distributional}.

\subsection{GFlowNets training criteria}

The objective of training GFlowNets is to enable the model to sample objects $\mathbf{x}$ 
with a probability proportional to $R(\mathbf{x})$. To achieve this goal, we introduce
several training criteria.

\paragraph{Flow matching (FM).}
We define a flow to be \textit{consistent} if it satisfies the flow matching constraint
for all internal states $\mathbf{s}$, meaning that the incoming flows equal the
outgoing flows: $\sum_{\mathbf{s}''\to\mathbf{s}}F(\mathbf{s}''\to\mathbf{s})=F(\mathbf
{s})=\sum_{\mathbf{s}\to\mathbf{s}'}F(\mathbf{s}\to\mathbf{s}')$. \citet{gflownet}
propose approximating the edge flow with a model $F_\theta(\mathbf{s},\mathbf{s}')$
parameterized by $\theta$ using the FM objective. For non-terminal states, the
objective is defined as $\mathcal{L}_\mathrm{FM}(\mathbf{s})=(\log\sum_{(\mathbf{s}''
\to\mathbf{s})\in\mathcal{A}}F_\theta(\mathbf{s}'',\mathbf{s})-\log\sum_{(\mathbf{s}
\to\mathbf{s}')\in\mathcal{A}}F_\theta(\mathbf{s},\mathbf{s}'))^2$. At terminal
states, a similar objective encourages the incoming flow to match the corresponding
reward. The objective is optimized using trajectories sampled from a training policy
$\pi$ with full support, such as a tempered version of $P_{F_\theta}$ or a mixture of
$P_{F_\theta}$ with a uniform policy $U$: $\pi_\theta=(1-\varepsilon)P_{F_\theta}+
\varepsilon U$. This approach is similar to $\varepsilon$-greedy and
entropy-regularized strategies in reinforcement learning (RL) to improve exploration.
\citet{gflownet} prove that if we reach a global minimum of the expected loss function
and the training policy $\pi_\theta$ has full support, then GFlowNet samples from the
target distribution.

\paragraph{Detailed balance (DB).}
The DB objective was proposed by \citet{foundations} to eliminate the need for
computationally expensive summing operations over the parents or children of states.
In DB-based learning, we train a neural network with a state flow model $F_\theta$, a
forward policy model $P_{F_\theta}(\cdot|\mathbf{s})$, and a backward policy model
$P_{B_\theta}(\cdot|\mathbf{s})$ parameterized by $\theta$. The optimization objective
is to minimize $\mathcal{L}_\mathrm{DB}(\mathbf{s},\mathbf{s}')=\left(\log(F_\theta(
\mathbf{s})P_{F_\theta}(\mathbf{s}'|\mathbf{s}))-\log(F_\theta(\mathbf{s}')P_{B_
\theta}(\mathbf{s}|\mathbf{s}'))\right)^2$. Sampling from the target distribution is
also done if a global minimum of the expected loss is achieved and $\pi_\theta$ has
full support.

\paragraph{Trajectory balance (TB).}
The TB objective, proposed by \citet{tb}, aims to enable faster credit assignment and
learning over longer trajectories. The loss function for TB is $\mathcal{L}_\mathrm{TB}
(\tau)=(\log(Z_\theta\prod_{t=0}^{n-1}P_{F_\theta}(\mathbf{s}_{t+1}|\mathbf{s}_t))-
\log(R(\mathbf{x})\prod_{t=0}^{n-1}P_{B_\theta}(\mathbf{s}_t|\mathbf{s}_{t+1})))^2$,
where $Z_\theta$ is a learnable parameter.

\section{Method}

We propose to enforce invariance constraints into the design of GFlowNets.
Here, invariance of a function $f$ under a group of transformations $G$ is defined
as $f(\mathbf{s})=f(g\cdot\mathbf{s})$, where group element $g$ acts on the input
domain of $f$. This means that inputs in the same orbit of $G$ are equivalent,
in the sense that they produce the same output when fed to $f$.

\subsection{State invariance}

Now consider the case where the state space of GFlowNet $\mathcal{S}$ has a
symmetric structure. Ideally we would require the forward policy of GFlowNet to
satisfy \textit{state invariance}:
\[
    P_F(\mathbf{s}'|\mathbf{s})=P_F(g\cdot\mathbf{s}'|g\cdot\mathbf{s}),\quad
    \forall\mathbf{s},\mathbf{s}'\in\mathcal{S},\forall g\in G,
\]
where $g$ is a transformation, such as a reflection in the $\mathcal{S}$ space.
If this constraint could be achieved, then the GFlowNet would obtain improved
performance since it will not waste capacity on unnecessary
modeling. The same idea applies to flows, edge flows, and backward policies:
\begin{align*}
    F(\mathbf{s}\to\mathbf{s}')=F(g\cdot\mathbf{s}\to g\cdot\mathbf{s}'),\ 
    F(\mathbf{s})=F(g\cdot\mathbf{s}),\\
    P_B(\mathbf{s}|\mathbf{s}')=P_B(g\cdot\mathbf{s}|g\cdot\mathbf{s}'),\quad
    \forall\mathbf{s},\mathbf{s}'\in\mathcal{S},\forall g\in G.
\end{align*}

\paragraph{Symmetrization via group averaging}
When training a GFlowNet with flow matching, we parameterize the edge flow function
$F(\mathbf{s}\to\mathbf{s}')$. One way to make the edge flow \emph{output} correct,
is to combine the output for equivalent actions, such as with averaging:
\[
    F(\mathbf{s}\to\mathbf{s}')=\frac1{|G|}\sum_{g\in G}\tilde{F}(g\cdot\mathbf{s}
    \to g\cdot\mathbf{s}').
\]
We can use a similar strategy to deal with the forward and backward policy, which are
needed in the detailed balance or trajectory balance parameterizations. For the state flow
function $F(\mathbf{s})$, we can set $F(\mathbf{s})=\frac1{|G|}\sum_{g\in G}\tilde
{F}(g\cdot\mathbf{s})$ for an arbitrary neural network $\tilde{F}(\cdot)$. The
same operation can be applied to the forward and backward policy modeling.
The disadvantage of such procedure is a potentially high computational cost when
the group $G$ is large, since enumerating all the group elements requires $\mathcal
{O}(|G|)$ complexity.

\paragraph{Symmetrization via canonical representation}
For some of the objects' representations, there is a more efficient way to detect
symmetry. If there exists a function $\mathcal{C}(\cdot)$ (we call $\mathcal{C}(
\mathbf{s})$ the \textit{canonical form} of $\mathbf{s}$) such that
\[
    \mathcal{C}(\mathbf{s})=\mathcal{C}(g\cdot\mathbf{s}),\quad\forall g\in G,
\]
then we could achieve state invariance with an arbitrary neural network $\tilde{F}$
as follows:
\[
    F(\mathbf{s}\to\mathbf{s}')\coloneqq\tilde{F}\bigl(\mathcal{C}(\mathbf{s})\to
    \mathcal{C}(\mathbf{s}')\bigr).
\]
The previous method needs $|G|$ forward passes of the neural network, while this
method only requires one forward pass. The disadvantage of this procedure is that
not all kinds of data possess a proper canonical form.

\subsection{Action invariance}

\begin{wrapfigure}{R}{0.5\textwidth}
    \centering
    \def\svgwidth{58mm}
    %% Creator: Inkscape 1.0 (4035a4fb49, 2020-05-01), www.inkscape.org
%% Accompanies image file 'action-symmetry.pdf' (pdf, eps, ps)
\begingroup%
  \makeatletter%
  \providecommand\color[2][]{%
    \errmessage{(Inkscape) Color is used for the text in Inkscape, but the package 'color.sty' is not loaded}%
    \renewcommand\color[2][]{}%
  }%
  \providecommand\transparent[1]{%
    \errmessage{(Inkscape) Transparency is used (non-zero) for the text in Inkscape, but the package 'transparent.sty' is not loaded}%
    \renewcommand\transparent[1]{}%
  }%
  \providecommand\rotatebox[2]{#2}%
  \newcommand*\fsize{\dimexpr\f@size pt\relax}%
  \newcommand*\lineheight[1]{\fontsize{\fsize}{#1\fsize}\selectfont}%
  \ifx\svgwidth\undefined%
    \setlength{\unitlength}{272.12598425bp}%
    \ifx\svgscale\undefined%
      \relax%
    \else%
      \setlength{\unitlength}{\unitlength * \real{\svgscale}}%
    \fi%
  \else%
    \setlength{\unitlength}{\svgwidth}%
  \fi%
  \global\let\svgwidth\undefined%
  \global\let\svgscale\undefined%
  \makeatother%
  \begin{picture}(1,0.54166667)%
    \lineheight{1}%
    \setlength\tabcolsep{0pt}%
    \put(0,0){\includegraphics[width=\unitlength,page=1]{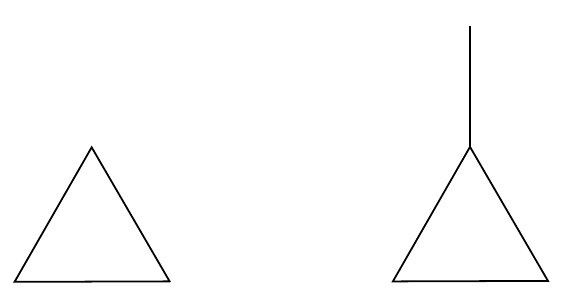}}%
    \put(0.15381238,0.28969276){\makebox(0,0)[lt]{\lineheight{1.25}\smash{\begin{tabular}[t]{l}1\end{tabular}}}}%
    \put(-0.01023684,0.01368133){\makebox(0,0)[lt]{\lineheight{1.25}\smash{\begin{tabular}[t]{l}2\end{tabular}}}}%
    \put(0.30275361,0.01387044){\makebox(0,0)[lt]{\lineheight{1.25}\smash{\begin{tabular}[t]{l}3\end{tabular}}}}%
    \put(0.84086283,0.28260577){\makebox(0,0)[lt]{\lineheight{1.25}\smash{\begin{tabular}[t]{l}1\end{tabular}}}}%
    \put(0.65476489,0.0148625){\makebox(0,0)[lt]{\lineheight{1.25}\smash{\begin{tabular}[t]{l}2\end{tabular}}}}%
    \put(0.9744486,0.01702027){\makebox(0,0)[lt]{\lineheight{1.25}\smash{\begin{tabular}[t]{l}3\end{tabular}}}}%
    \put(0.83312233,0.50308156){\makebox(0,0)[lt]{\lineheight{1.25}\smash{\begin{tabular}[t]{l}4\end{tabular}}}}%
    \put(0,0){\includegraphics[width=\unitlength,page=2]{action-symmetry.pdf}}%
  \end{picture}%
\endgroup%

    \caption{An example of action symmetry. There are three possible actions leading
    from the left graph to the right graph, but only one action leading backwards.}
    \label{fig:action-symmetry}
    \vspace{-20pt}
\end{wrapfigure}

Symmetries may exist not only in the state space but also in the actions involved.
Let's consider the graph generation environment, where actions include adding a node
to an existing node and connecting them with a new edge. In
Figure~\ref{fig:action-symmetry}, we can see that there are three possible actions
that lead from the left graph to the right graph. These actions involve adding a new
node to any of the three existing nodes. We refer to them as \textit{isomorphic
actions} because they all result in the same (\textit{i.e.}, isomorphic) graph. However, in the
opposite direction, there is only one action that leads from the right graph to the
left graph, which is the removal of node 4 and the edge connecting node 1 and node 4.
Currently, GFlowNet training pipelines do not consider such isomorphic actions and would treat these three isomorphic actions as distinct, resulting
in an incorrect forward edge flow that is three times larger than it should be.

In order to model the correct flow, it is necessary to identify symmetric actions
and sum their probabilities during the training process. Since these symmetric actions
do not have an efficient canonical form, it is essential to enumerate all of these
actions given a graph-action pair.

\paragraph{Symmetrization via isomorphism testing\hfill}
One straightforward method to enumerate isomorphic actions of a given action is to
iterate through all possible actions and verify if they generate isomorphic graphs.
However, performing direct isomorphism checks is computationally expensive and requires
a time complexity of $\mathcal{O}(n\times n!)$, as described by \citet{vf2}, which
makes it impractical for larger graphs. Hence, in practice, we require approximation
schemes to overcome this limitation.

\paragraph{Symmetrization via graph-level positional encoding}
An alternative approach that improves efficiency over direct isomorphism testing
involves using graph-level positional encoding (PE). PE functions as an embedding
technique that maps a graph to a representation vector. Isomorphic graphs are
guaranteed to have identical graph-level PEs. However, due to the inherent limitations
in the expressive power of PE, it is possible for two non-isomorphic graphs to share
the same graph-level PE. Our objective is to use graph-level PEs that are both
expressive and computationally efficient, meaning they can effectively differentiate
between most non-isomorphic graphs.

We investigate three kinds of PEs: PEs produced by the 1-WL test \citep{wl}, random
walk positional encoding (RWPE) \citep{distance-encoding}, and sum of edge features.
\begin{enumerate}
    \item The 1-WL test is a color refinement method that finds for each node in each
    graph a signature based on the neighborhood around the node. These signatures can
    then be used to find the correspondance between nodes in the two graphs, which can
    be used to check for isomorphism.
    \item The RWPE is defined as the concatenation of the diagonal elements of powers
    of the random walk matrix: $\vx_i=[(\mA\mD^{-1})_{ii}^k],k=1,2,\dots$, where
    $\vx_i$ is the RWPE of node $i$ and $\mA\mD^{-1}$ is the random walk matrix of
    the graph. The original RWPE does not consider node colors, thus it cannot
    distinguish graphs with the same structure but different node colors. Here we
    propose to incorporate node colors into RWPE, by multiplying the powers of the
    random walk matrix with node colors: $\vx_i=[(\vc^\mathrm{T}\mA\mD^{-1})_i^k],
    k=1,2,\dots$, where $\vc$ is a vector representing node colors. We then take
    $\sum_i\vx_i$ as the graph-level PE.
    \item We could also calculate all edge features by summing\footnote{In practice
    we take the sum of squares instead of sum to avoid collision.} node features of
    their vertices and considering their sum as the graph-level feature. That is,
    for edge $(i,j)$, its edge feature is defined as $\ve_{ij}\coloneqq\vx_i+\vx_j$
    and we could take $\sum_{(i,j)\in\sE}\ve_{ij}$ as the graph-level PE.
\end{enumerate}

We evaluate these three PE methods on graphs with a maximum of 7 nodes, where each
node can be assigned one of two available colors. The actions considered in these
graphs are the addition of nodes (\textit{i.e.}, adding a node to an existing node and
connecting them with an edge) and the addition of edges (\textit{i.e.}, introducing a
previously non-existent edge). We systematically test all possible combinations of
graph-action pairs and assess the PE methods' ability to accurately enumerate all
isomorphic actions. We record the error rate and running time for each PE method,
and the results are summarized in Table~\ref{tab:graph-level-pe}.

\begin{table}[htbp]
    \centering
    \caption{The running time and error rate of different PE methods, where ``WL''
    represents the 1-WL test feature, ``RW'' refers to random walk positional
    encodings (multiplied with node colors), and ``edge'' signifies the calculation
    of all edge features (by summing node features of their vertices) and
    considering their sum as the graph-level feature.}
    \begin{tabular}{ccc}
        \toprule
        PE method      & Running time   & Error rate \\
        \midrule
        WL             & 7h17min        & 989650/2483411 = 0.3985 \\
        WL + edge      & 7h51min        & 485123/2483411 = 0.1953 \\
        RW             & 4h24min        & 1493301/2483411 = 0.6013 \\
        RW + edge      & 4h41min        & 38/2483411 = 0.000013691 \\
        WL + RW        & 10h59min       & 952227/2483411 = 0.3834 \\
        WL + RW + edge & 10h56min       & 0/2483411 = 0 \\
        \bottomrule
    \end{tabular}
    \label{tab:graph-level-pe}
\end{table}

As shown in Table~\ref{tab:graph-level-pe}, both ``RW + edge'' and ``WL + RW + edge''
achieves almost perfect accuracy in discovering isomorphic actions. However the former
is much faster to compute than the latter, thus we use ``RW + edge'' as PE in our
experiments.

\paragraph{Symmetrization via node-level and edge-level positional encoding}
Symmetrization via graph-level PE requires iterating through all possible actions
and calculating the graph-level PEs of each preceding graph of these actions, which
could be inefficient since we only care about actions on the single original graph.
Here we propose an approximating alternative, by calculating the node-level and
edge-level PEs of the original graph. If we wish to verify whether two node-adding
actions are isomorphic, we can calculate the node-level PEs of their corresponding
nodes and check whether they are identical. Similarly, we verify whether two
edge-adding actions are isomorphic by checking whether the edge-level PEs of their
corresponding edges are identical. In this way, we only need to calculate the PEs of
nodes and edges of the original graph, which is more efficient than calculating the
graph-level PEs of all preceding graphs. The node-level and edge-level PEs are the
same as described above, without taking their sum as graph-level feature.

\begin{wrapfigure}{R}{0.65\textwidth}
    \vspace{-15pt}
    \centering
    \def\svgwidth{80mm}
    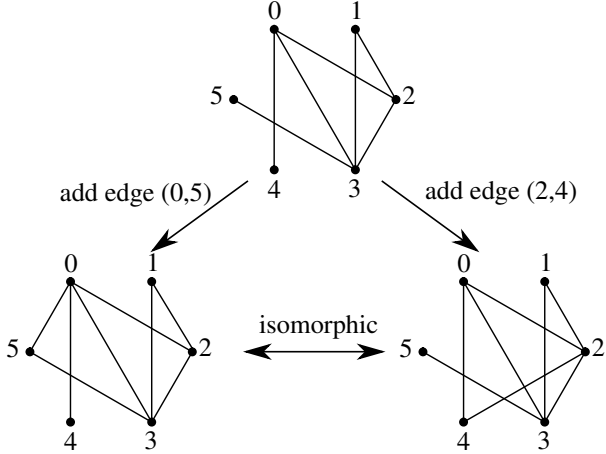
    \caption{An example where two actions with different edge-level PEs could lead
    to isomorphic graphs.}
    \label{fig:counterexample}
    \vspace{-20pt}
\end{wrapfigure}

One disadvantage of using node-level and edge-level PEs is that actions with distinct
PEs do not always lead to non-isomorphic graphs. We give a counterexample in
Figure~\ref{fig:counterexample}. As shown in Figure~\ref{fig:counterexample},
edge (0,5) and edge (2,4) are structurally dissimilar, thus they have different PEs,
yet they lead to isomorphic graphs. Luckily such counterexamples are quite rare
(1 in a few thousand graphs) and they have negligible impact on model's performance.
We use node-level and edge-level PEs in our experiments.

\section{Experiments}

\subsection{Experiments on state invariance}

\begin{wrapfigure}[12]{r}{0.25\textwidth}
    \vspace{-20pt}
    \begin{center}
    \includegraphics[width=0.22\textwidth]{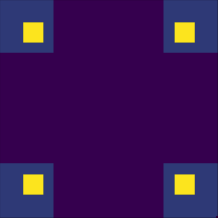}
    \end{center}
    \caption{The HyperGrid environment.}
    \label{fig:hypergrid}
\end{wrapfigure}

We implement and evaluate our method on the benchmark adopted by \citet{gflownet,
tb}, namely hypergrid exploration sampling tasks. Starting from one corner, an
agent moves in a grid-like world to explore the landscape defined by the following
reward function in Equation~\eqref{eq:hypergrid}. The agent starts at the fixed
down left corner in every episode, and is only allowed to move up or right at each
step. A third stop action indicates to terminate the trajectory and leave the agent
at the grid it stands. Ideally, the agent should learn to stop at relatively high
reward regions at terminating times. Also, we do not want the agent's solution to
converge to a particular mode, but would like it to discover all different modes
(4 modes in example in Figure~\ref{fig:hypergrid}). Therefore, we use the L1
distance between the empirical distribution (sampled by the GFlowNet agent) and
the ground truth distribution as the evaluation metric.
\begin{equation}\label{eq:hypergrid}
    R(\vx)=R_0+0.5\prod_{d=1}^D\mathrm{I}[0.25<|\evx_d-0.5|]+2\prod_{d=1}^D
    \mathrm{I}[0.3<|\evx_d-0.5|<0.4].
\end{equation}

In Equation~\eqref{eq:hypergrid}, $\vx=(\evx_1,\dots,\evx_D)$ where $D$ is the
number of state dimension. Notice that the reward function is invariant to the
permutation of the coordinate ordering. This indicates that our flow model should
be invariant to the permutation of state coordinates, \textit{i.e.}, the group
$G$ contains all permutations of $D$ elements. Notice that under such scenario,
the proposed enumeration based method would require $D!$ times of neural network
forward pass. On the other hand, in order to conduct the second canonization
based method, we define
\[
    \mathcal{C}(\vx)\coloneqq(\evx_{\pi(1)},\dots,\evx_{\pi(D)}),\quad
    \evx_{\pi(1)}\leq\dots\leq\evx_{\pi(D)},
\]
where $\pi$ denotes the permutation to arrange the coordinates in descending order.
Notice that on the states which contains two equal coordinates, our method cannot
guarantee exact invariance and thus being an approximation (yet efficient) method.

\begin{figure}[htbp]
    \centering
    \includegraphics[scale=0.55]{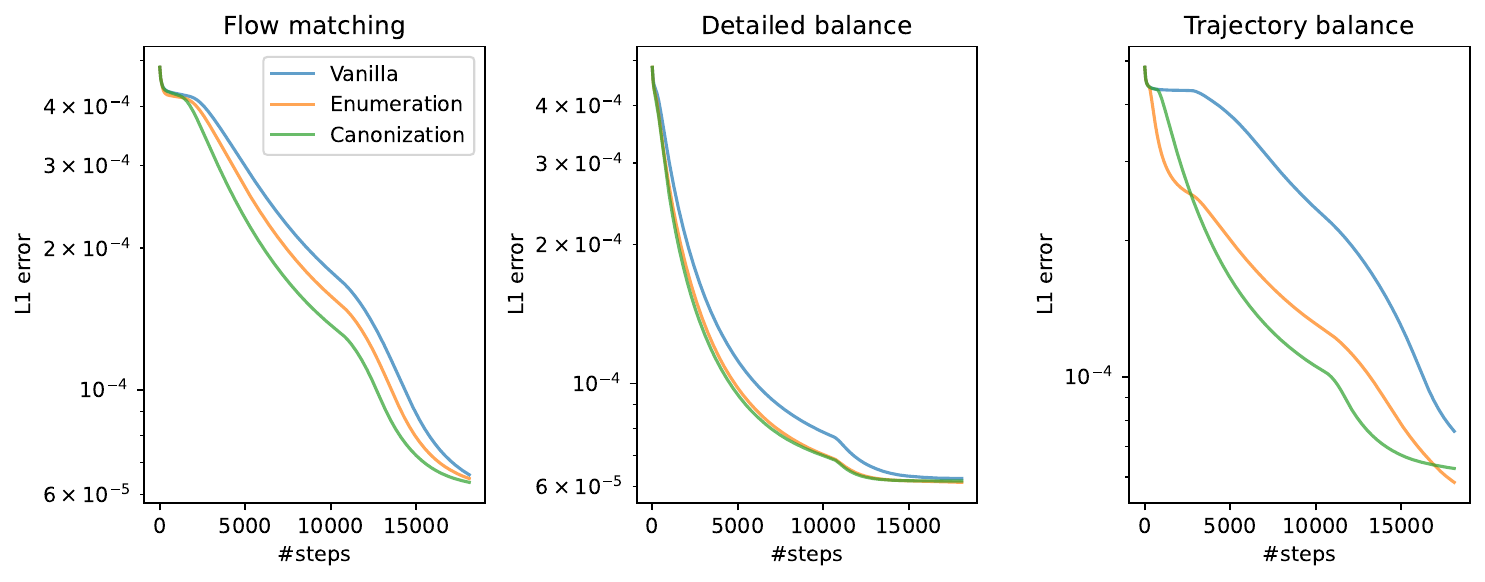}
    \caption{Results of the baseline, our method 1 (enumration based), and our
    method 2 (canonization based) on the hypergrid environment. We test with
    $\mathrm{horizon}=16$, $\mathrm{dimension}=3$, $R_0=0.001$.}
    \label{fig:hypergrid-results}
\end{figure}

We plot the L1 error with regards to the number of steps (\textit{i.e.}, the number
of visited states). The results in Figure~\ref{fig:hypergrid-results} indicate that
our proposed methods consistently outperform the original baseline, showing a
better sample efficiency for symmetry involved methods.

\subsection{Experiments on action invariance}

To test the effects of action invariance, we setup a simple environment with all
possible graphs of a maximum of 7 nodes, where the nodes can be one of two colors.
We define three different reward functions with varying difficulty. The hardest
function, \textbf{cliques}, requires the model to identify subgraphs in the state
which are 4-cliques of at least 3 nodes of the same color. The second function,
\textbf{neighbors}, requires the model to verify whether nodes have an even number
of neighbors of the opposite color. Finally, the third function, \textbf{counting},
simply requires the model to count the number of nodes of each color in the state.

This environment has a total of 72296 states, which allows us to compute the
ground truth probability $p(x)$ and the learned probability $p_\theta(x)$ relatively
quickly. We compare three models, namely vanilla GFlowNet, GFlowNet with isomorphism
testing (ground truth), and GFlowNet with positional encodings. The JS divergence
between $p(x)$ and $p_\theta(x)$ are reported. We only report results on the counting
reward and defer full experiment results to Appendix~\ref{app:full-results}.

\begin{figure}[htbp]
    \centering
    \includegraphics[scale=0.55]{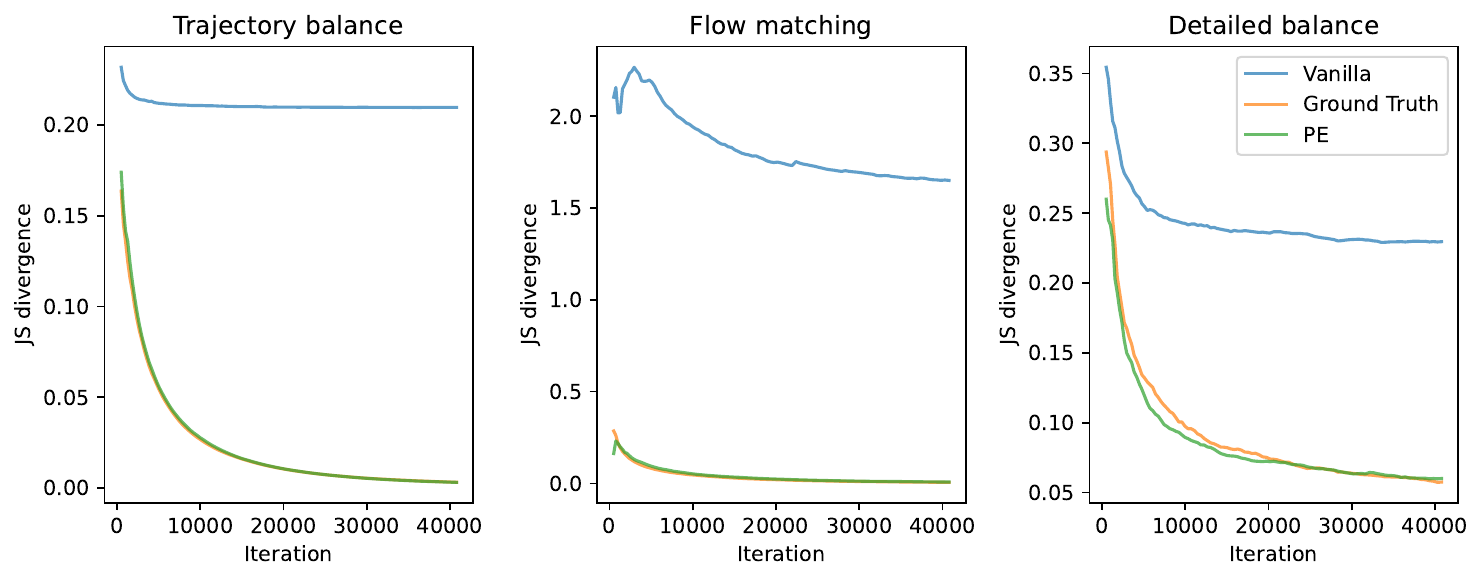}
    \caption{The JS divergence between $p(x)$ and $p_\theta(x)$ during offline
    GFlowNet training.}
    \label{fig:cliques-results}
\end{figure}

As shown in Figure~\ref{fig:cliques-results}, incorporating action invariance
improves the accuracy of the learned flow functions. Direct isomorphism testing and
positional encodings achieve comparable performance, but the former has a time
complexity of $\mathcal{O}(n\times n!)$ while the latter has a time complexity of
$\mathcal{O}(n^3)$.

\begin{wrapfigure}{R}{0.5\textwidth}
    \vspace{-15pt}
    \centering
    \includegraphics[width=0.48\textwidth]{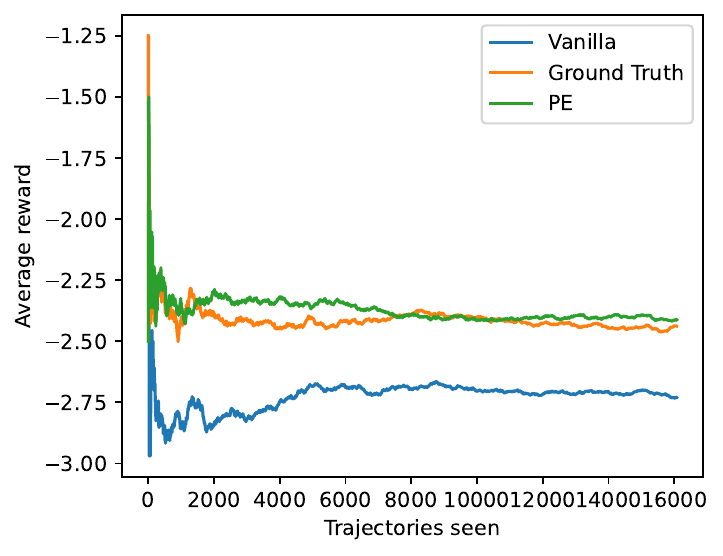}
    \caption{The average reward during training.}
    \label{fig:cliques-reward}
    \vspace{-40pt}
\end{wrapfigure}

We also report the average reward during training in Figure~\ref{fig:cliques-reward}.
As shown in Figure~\ref{fig:cliques-reward}, both direct isomorphism testing and
positional encodings achieve better performance than vanilla GFlowNet, indicating
more accurate flow probabilities indeed lead to higher rewards.

\section{Conclusion}

In this paper we propose to incorporate invariance to the internal symmetries within
the generation process into GFlowNet training. For symmetric states, we propose
enumeration-based and canonization-based symmetrization to achieve state invariance.
For symmetric actions, we propose direct isomorphism testing and positional encodings
as an efficient alternative. Results on synthetic experiments validate the efficacy
of our proposed approaches.

\bibliographystyle{plainnat}
\bibliography{references.bib}

\newpage

\appendix

\section{Full experiment results}\label{app:full-results}

\begin{figure}[htbp]
    \centering
    \includegraphics[scale=0.55]{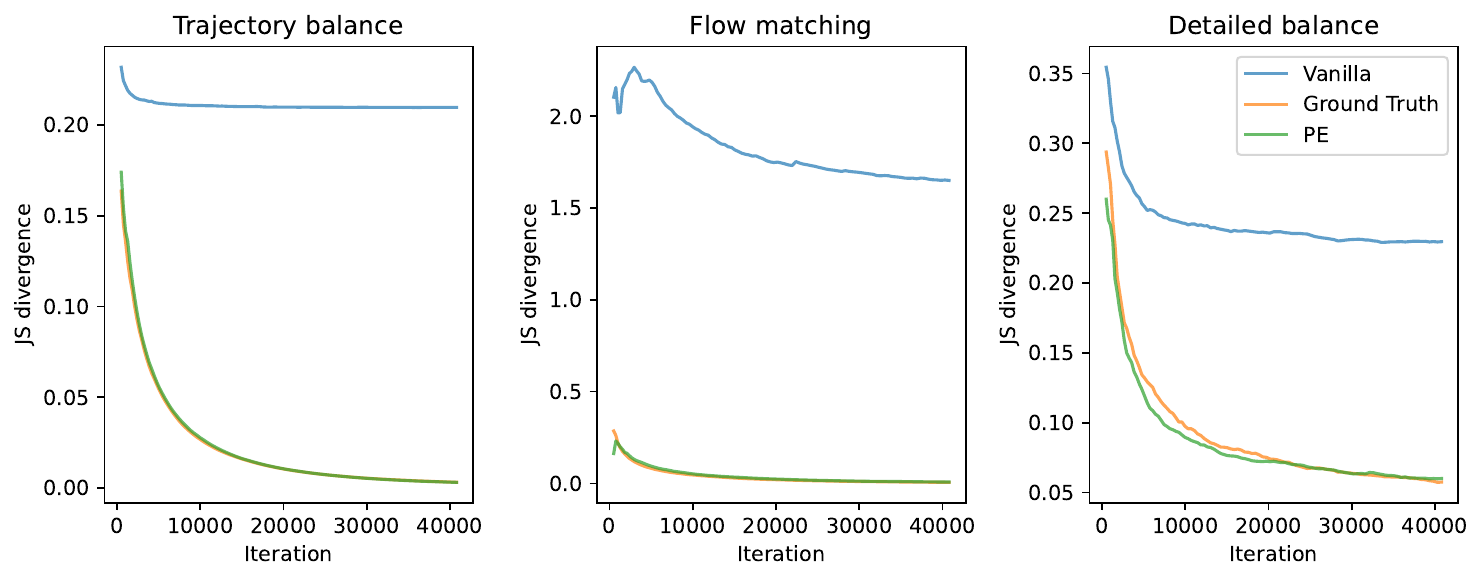}
    \caption{Results on the counting reward function.}
    \label{fig:counting}
\end{figure}

Results on the \textbf{counting} reward function are reported in
Figure~\ref{fig:counting}.

\begin{figure}[htbp]
    \centering
    \includegraphics[scale=0.55]{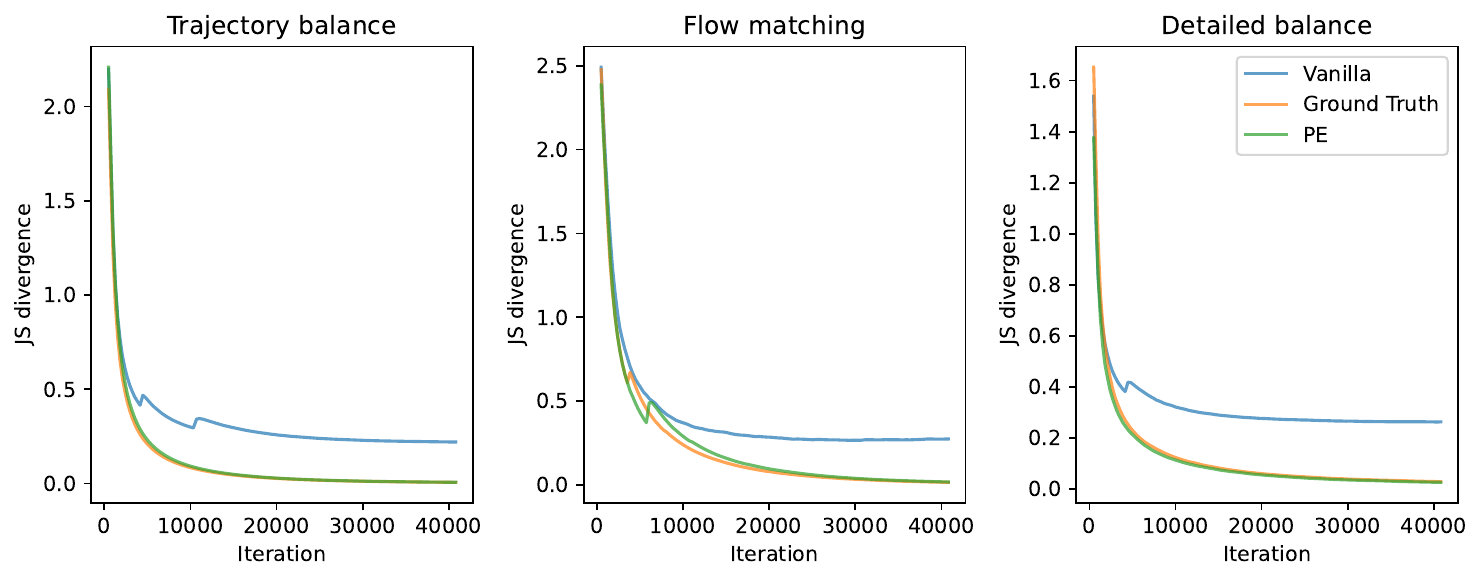}
    \caption{Results on the neighbors reward function.}
    \label{fig:neighbors}
\end{figure}

Results on the \textbf{neighbors} reward function are reported in
Figure~\ref{fig:neighbors}.

\begin{figure}[htbp]
    \centering
    \includegraphics[scale=0.55]{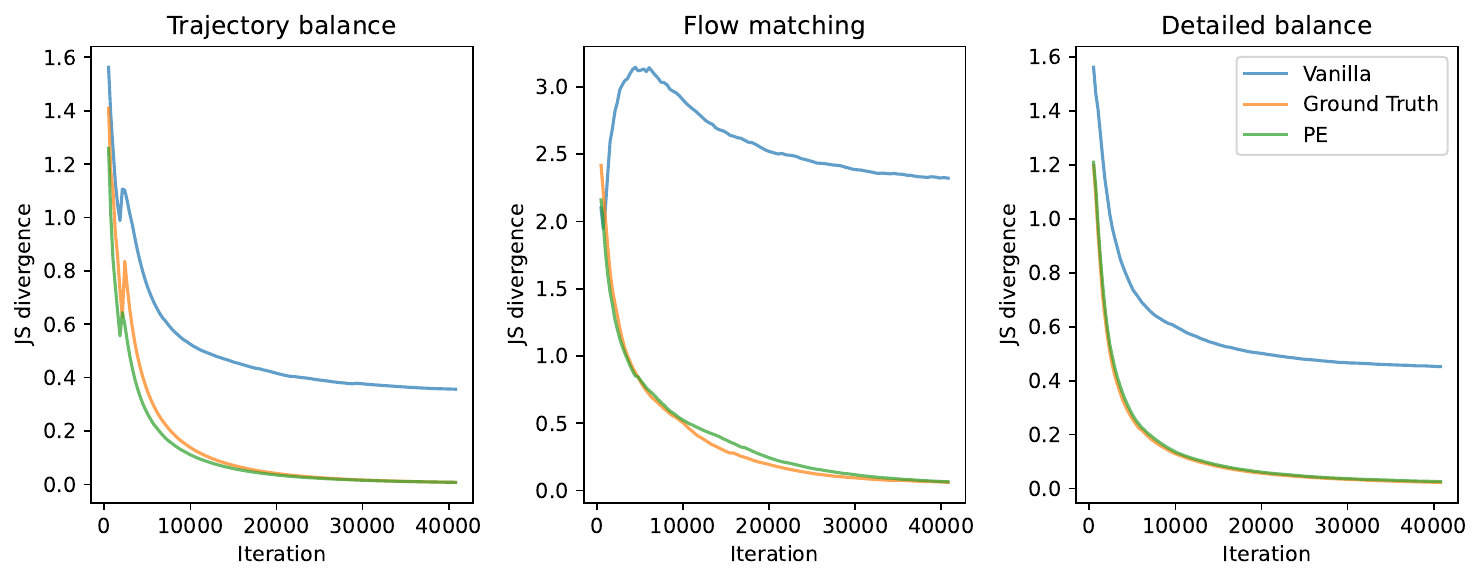}
    \caption{Results on the cliques reward function.}
    \label{fig:cliques}
\end{figure}

Results on the \textbf{cliques} reward function are reported in
Figure~\ref{fig:cliques}.

\end{document}